\documentclass[letterpaper, 10 pt, conference]{ieeeconf}  
\IEEEoverridecommandlockouts                              
\usepackage{xcolor}
\usepackage{graphicx}
\usepackage{gensymb}
\usepackage[hyphens]{url}
\usepackage{multirow}
\usepackage{multicol}
\usepackage{tabularx}
\usepackage{dsfont}
\usepackage{url}
\usepackage{color}
\usepackage{subcaption}
\usepackage{caption}
\usepackage{booktabs}
\usepackage{dcolumn}
\usepackage{amssymb}
\usepackage{hyperref}
\usepackage{amsmath}
\usepackage{amsfonts}
\usepackage{colortbl}
\usepackage{tikz}
\colorlet{lightgray}{gray!20}

\hypersetup{
    colorlinks=false,
    linkcolor=blue,
    filecolor=magenta,      
    urlcolor=cyan,
    pdftitle={Overleaf Example},
    pdfpagemode=FullScreen,
    }

\title{\LARGE \bf
V2VSSC: A 3D Semantic Scene Completion Benchmark for Perception with Vehicle to Vehicle Communication}

\author{Yuanfang Zhang$^{\ast}$$^{\dagger}$, Junxuan Li$^{\ast}$, Kaiqing Luo, Yiying Yang, Jiayi Han, 
\\Nian Liu, Denghui Qin, Peng Han, Chengpei Xu
\thanks{$^{\ast}$Equal contribution}
\thanks{$^{\dagger}$Corresponding Author}
\thanks {Yuanfang Zhang is with Nanjing University of Information Science and Technology, School of Computer Science. and Autocity (Shenzhen) Autonomous Driving Co.,ltd,{\{\tt\small{zyf.robinzhang}\}}{\tt\small{@gmail.com}}}%
\thanks{Junxuan li is with South China Normal University, School of Electronics and Information Engineering,
{\{\tt\small{junxuanli}\}}{\tt\small{@m.scnu.edu.cn}}}%
\thanks{Kaiqing Luo and Peng Han are with South China Normal University, School of Electronics and Information Engineering,
{\{\tt\small{kqluo, hanp}\}}{\tt\small{@scnu.edu.cn}}}%
\thanks{Yiying Yang is with Endeavor Star Co.,ltd, Australia,
{\{\tt\small{yyiying68}\}}{\tt\small{@163.com}}}%
\thanks{Jiayi Han is with Inspur Group,
{\{\tt\small{hanjiayi}\}}{\tt\small{@inspur.com}}}%
\thanks{Nian Liu is with Mohamed bin Zayed University of Artificial Intelligence, Department of Computer Vision,
{\{\tt\small{liunian228}\}}{\tt\small{@gmail.com}}}%
\thanks{Denghui Qin is with Peking University, College of Engineering,
{\{\tt\small{qindenghui}\}}{\tt\small{@pku.edu.cn}}}%
\thanks{Chengpei Xu is with University of New South Wales, Faculty of Engineering,
{\{\tt\small{Chengpei.Xu}\}}{\tt\small{@unsw.edu.au}}}%
}

\begin{document}
\maketitle
\pagestyle{empty} 
\thispagestyle{empty} 

\begin{abstract}
Semantic scene completion (SSC) has recently gained popularity because it can provide both semantic and geometric information that can be used directly for autonomous vehicle navigation. However, there are still challenges to overcome. SSC is often hampered by occlusion and short-range perception due to sensor limitations, which can pose safety risks. This paper proposes a fundamental solution to this problem by leveraging vehicle-to-vehicle (V2V) communication. We propose the first generalized collaborative SSC framework that allows autonomous vehicles to share sensing information from different sensor views to jointly perform SSC tasks. To validate the proposed framework, we further build V2VSSC, the first V2V SSC benchmark, on top of the large-scale V2V perception dataset OPV2V. Extensive experiments demonstrate that by leveraging V2V communication, the SSC performance can be increased by 8.3\% on geometric metric IoU and 6.0\% mIOU.
\end{abstract}

\section{INTRODUCTION}
Holistic 3D scene understanding is a critical component of autonomous driving. Autonomous vehicles (AVs) rely on precise 3D geometry and semantics to safely navigate around obstacles. To achieve this, many researchers are investigating the use of bird's-eye view (BEV) maps to represent the surrounding information~\cite{philion2020lift, huang2021bevdet, li2023bevdepth, liang2022bevfusion}. While BEV maps can efficiently represent the surrounding information, they compress height information and limit the granularity of the navigation map, which could potentially cause safety issues, especially in locations where altitude information is critical.

In response, recent research has turned its focus to 3D Semantic Scene Completion (SSC). The goal of SSC is to predict full 3D voxel details, including occupancy and semantics, even when only given partial observations. This mirrors human perception: often, we only see part of a 3D scene but can infer the rest. Due to its similarity to human perception and its potential to provide more driving insights, SSC is becoming an increasingly researched topic in the field.

A fundamental challenge in SSC is extracting complete 3D data from only partial observations. This often requires models to overcome occlusions and achieve a broad perceptual range. Existing solutions like the VoxFormer~\cite{li2023voxformer} utilize a combination of vision transformer~\cite{dosovitskiy2020image} and masked auto-encoder~\cite{he2022masked} for this purpose. On the other hand, TPVFormer~\cite{huang2023tri} uses a triple-plane transformer to tackle occlusions. However, it's crucial to note that the occlusion problem is intrinsically complex due to the limitations of single-view sensors. Depending solely on deep learning, no matter how advanced, might not be the definitive solution.

\begin{figure*}
    \centering
    \includegraphics[width=0.9\linewidth]{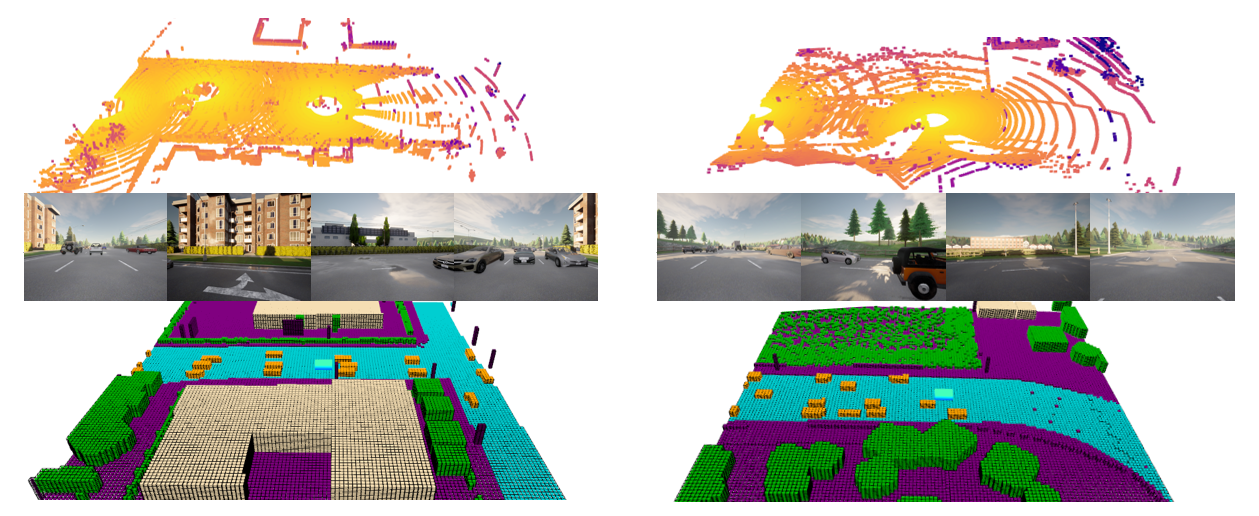}
    \caption{\textbf{Two samples from Our V2VSSC.} \emph{Top}: The aggregated point cloud from surrounding CAVs. \emph{Middle:} 4 views camera of the ego vehicle. \emph{Down:} The semantic occupancy map generated by the ego vehicle.}
    \label{fig:Samples}
\end{figure*}

In our research, we pivot to a different approach for SSC by tapping into Vehicle-to-Vehicle (V2V) communication technology. This technology enables adjacent AVs to share sensing information, thereby offering diverse viewpoints of a single scene, potentially bypassing occlusion challenges. Most existing V2V studies have been centered around 3D object detection~\cite{xu2022v2xvit, wang2020v2vnet} and BEV semantic segmentation~\cite{xu2022cobevt}, leaving a gap in the SSC realm. To bridge this, we leverage the popular V2V perception dataset, OPV2V~\cite{xu2022opv2v}, to create the inaugural V2V SSC benchmark. Figure~\ref{fig:Samples} demonstrates some examples that are generated in our benchmark. This benchmark features four distinct models and encompasses six semantic categories, including road, cars, terrain, building, vegetation, and poles. Extensive experiments in the benchmark demonstrate that by utilizing V2V communication technology, the SSC geometric Interest-of-Union~(IoU) can increase by 8.3\%, and the semantic metric mIoU can rise 6\%. Our contribution can be summarized as follows:
\begin{itemize}
    \item  We introduce a novel approach to 3D Semantic Scene Completion (SSC) by leveraging Vehicle-to-Vehicle (V2V) communication, allowing autonomous vehicles to share sensing information and provide multiple perspectives on a single scene, effectively addressing occlusion challenges.
    \item We establish the first V2V SSC benchmark, based on the renowned OPV2V perception dataset. This benchmark comprises four unique models and spans six semantic categories, setting a standard for future research in this direction.
    \item Through extensive experiments, we demonstrate that the use of V2V communication in SSC leads to significant improvements in metrics: an 8.3\% boost in geometric IoU and a 6\% rise in the semantic metric mIoU, underscoring the efficacy and potential of our approach in real-world scenarios.
\end{itemize}

\section{Related Work}
\noindent\textbf{Camera-based Semantic Scene Completion:} SSCNet~\cite{song2017semantic} initially defined the SSC task, emphasizing the mutual benefits of semantic segmentation and scene completion. Recognizing object fragments aids in estimating its 3D positioning, while the knowledge of 3D occupancy informs about an object's shape and size. However, SSCNet exclusively relies on depth maps, neglecting attributes such as color and texture. TS3D~\cite{garbade2019two} enhances SSCNet's methodology by implementing a dual-stream feature extractor, capturing richer information. AMFNet~\cite{li2020attention} capitalizes on established 2D semantic segmentation methods, using 2D segmentation information to concurrently guide 3D scene completion and semantic segmentation. MonoScene~\cite{cao2022monoscene} derives dense 3D rasterized semantic scenes from a solitary RGB image, bypassing the need for depth maps. VoxFormer~\cite{li2023voxformer} introduces a Transformer-based approach with a dual-phase design, starting with a depth estimator followed by dense voxel generation through a self-attention mechanism. OccDepth~\cite{miao2023occdepth} exploits the implicit depth information from stereo images for improved 3D geometric structures and superior 3D perceptual feature integration. TPVFormer~\cite{huang2023tri} advocates for a three-perspective representation and calculates the attention map across different views.

\noindent\textbf{LiDAR-based Semantic Scene Completion:}  There are also many works that focus on leveraging LiDAR sensors to complete semantic scenes. The accuracy of LiDAR in providing precise distance measurements combined with fine-grained semantic details makes it invaluable for 3D vision in autonomous driving, especially when compared to data from cameras and other sensors. LMSCNet~\cite{roldao2020lmscnet}, in response to the advent of the SemanticKITTI~\cite{behley2019semantickitti} dataset, crafts a lightweight architecture combining 2D and 3D convolutions, employing a 2D UNet backbone and a 3D segmentation header. S3CNet~\cite{cheng2021s3cnet} presents a sparse tensor-based methodology for predicting 3D scene completion and semantic segmentation from a singular unified LiDAR point cloud. JS3C~\cite{yan2021sparse} adopts contextual shape priors from continuous LiDAR sequences. It introduces the Point Voxel Interaction (PVI) module to strengthen the fusion of knowledge between Semantic Segmentation (SS) and SSC tasks, enhancing the interplay between local geometry and global voxel structure. In this paper, we mainly focus on collaborative LiDAR-based SSC.

\noindent\textbf{Multi-Agent Perception: } With advancements in communication technologies, coupled with the rise of V2V applications, collaborative perception has become popular. It studies how to fuse the sensing information from different agents to achieve better perception performance. OPV2V~\cite{xu2022opv2v} uses a single-head self-attention module for feature fusion, targeting streamlined processing. F-Cooper~\cite{chen2019f} deviates from traditional fusion techniques by utilizing the maxout fusion operation. V2VNet~\cite{wang2020v2vnet} introduces a spatially aware message-passing mechanism, emphasizing simultaneous detection and prediction tasks. DiscoNet~\cite{li2021learning} employs knowledge distillation to refine training. It restricts feature mapping to align with early fusion in the network, ensuring consistent knowledge transfer. CoBEVT~\cite{xu2022cobevt} proposed a fused axial attention module, focusing on capturing interactions both across diverse views and the broader agent global space, specifically in collaborative BEV semantic segmentation. Coca3D~\cite{hu2023collaboration} introduces multi-agent collaboration to eliminate the ambiguity of estimating the depth with only camera inputs. All of these previous works focus on the 3D object detection task, while in this paper, we shift the focus toward a new task -- SSC.

\section{Dataset Curation}
\subsection{Revisit of outdoor SSC dataset}
SemanticKITTI\cite{behley2019dataset}, established as the pioneering outdoor SSC benchmark, offers semantic annotations spanning 22 driving sequences centered in Karlsruhe, Germany. This dataset employs voxelization, a process that transforms unstructured point cloud data into coherent 3D representations. To produce ground truth labels, SemanticKITTI relies on point cloud registration across multiple LIDAR scans. However, this approach presents inherent challenges.

Contrasting with SemanticKITTI, our methodology harnesses the Carla simulator~\cite{Dosovitskiy17}. We procure precise occupancy and semantic details directly from the server. This approach ensures enhanced representation accuracy, eliminating the complications associated with point cloud registration.

\subsection{V2VSSC Dataset}
Our V2VSSC dataset is an extension of the groundbreaking OPV2V~\cite{xu2022opv2v}, which is a significant advancement in the domain of autonomous multi-vehicle collaboration research. The OPV2V dataset contains over 70 different scenarios, including urban roads, highways, and mountain roads. In each scenario, there exists a minimum of 2 connected vehicles and a maximum of 7, with an average of approximately 3. OPV2V provides detailed information on all objects, including the ID, position, heading angle, velocity, car model, and color. Therefore, we are able to reply the driving log in the CARLA simulator and retrieve the semantic occupancy map from the simulator server directly to construct our SSC benchmark. 

Creating a semantic occupancy map for each CAV is a multi-tiered process. Initially, the complete topology of lane lines is extracted using the CARLA simulator. Points surrounding the CAV, within a 70-meter radius, are then refined, converted to the CAV's coordinate frame, and subsequently voxelized. Following this, redundant point pairs are removed, with spatial interpolation techniques employed to sketch out the drivable space. The next phase involves incorporating all stationary elements: the terrain, architecture, vegetation, and poles. Each element's 3D boundaries and orientations are extracted from CARLA, which are then projected onto the CAV's frame and voxelized. This approach is mirrored for dynamic objects.

However, crafting the boundary of the semantic grid map presents a unique challenge, particularly when only portions of an object fit within the map. In such instances, a meticulous filtering process becomes essential, ensuring retention of only those points that align with the map's parameters while discarding extraneous data.

Despite obtaining voxel coordinates rich in varied semantic data, challenges persist. The simulator-generated bounding boxes sometimes overlap, creating issues in ensuring that semantic voxel coordinates remain distinct. To counteract this, we've prioritized specific categories: cars, drivable regions, poles, vegetation, buildings, and terrain. In instances of overlap, voxels are classified under the highest-priority category. Additionally, nuances arise with vegetation categorization. Due to the diversity of tree models in the simulator, it's challenging to consistently assign voxels. Given our focus on vehicular navigation, we've implemented a height-based division for trees, assigning lower-height voxels to trunks and upper ones to the canopy. Conclusively, every empty voxel in the occupancy map is furnished with distinct semantic data, yielding the final semantic occupancy map for the respective CAV.

\begin{figure}[!h]
    \centering
    \includegraphics[width=0.9\linewidth]{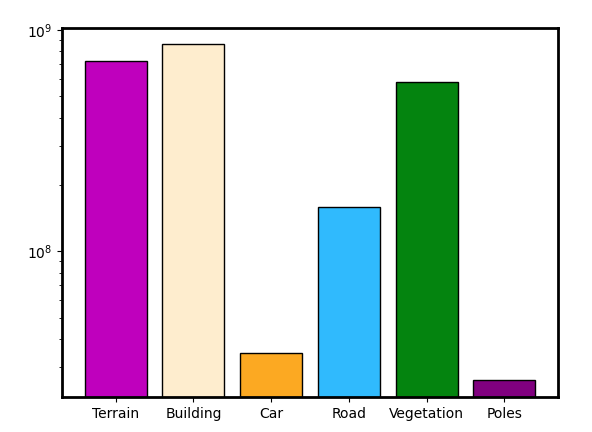}
    \caption{\textbf{Label Distribution.} X-axis represents the semantic category, Y-axis represents the number of voxels.}
    \label{fig:statistics}
\end{figure}

\begin{figure*}
    \centering
    \includegraphics[width=0.8\linewidth]{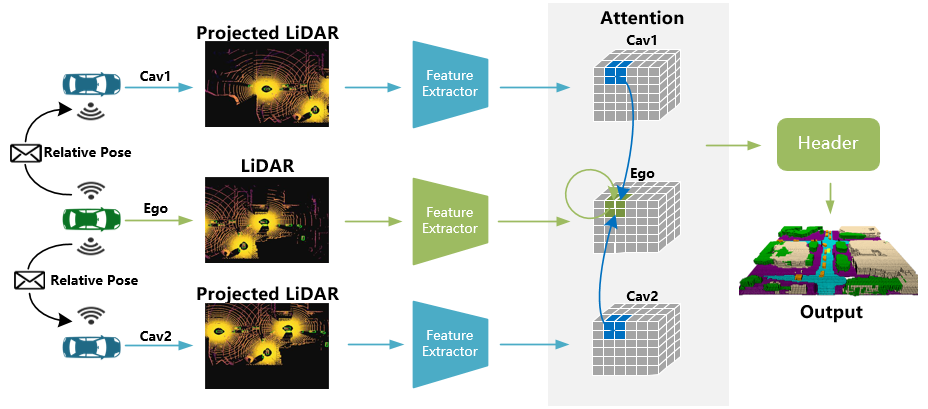}
    \caption{\textbf{The structure of Intermediate Fusion pipeline.} }
    \label{fig:system}
\end{figure*}

\begin{table*}[!t]
    \centering
    \def\xwidth{0.3}
    \caption{\textbf{Performance on V2VSSC dataset(test set).} We report the geometric metric Iou, semantic metric mIoU, critical objects iou cIoU and the IoU for each semantic class.}
    \begin{tabular*}{0.71\linewidth}{c|ccc|cccccc}
         \cellcolor{lightgray}{Method} &\cellcolor{lightgray}{IoU} & \cellcolor{lightgray}{mIoU} & \cellcolor{lightgray}{cIoU}  &\cellcolor{lightgray}{road} & \cellcolor{lightgray}{car} & \cellcolor{lightgray}{terrian} & \cellcolor{lightgray}{building} & \cellcolor{lightgray}{vegetation} & \cellcolor{lightgray}{poles} \\
        \toprule
         No Fusion & 53.2 & 42.2 & 60.0 & 65.6 & 54.4 & 48.3 & 31.8 & 41.2 & 12.0\\
         Early Fusion & 56.1 & 45.7 & \textbf{66.8} & 68.8 & \textbf{64.8} & 48.1 & 33.1 & 42.9 & 16.4\\
         Intermediate Fusion & 55.6 & 45.1 &64.6 & 66.8 & 62.4 & 48.3 & 33.1 & 43.3 & 16.7\\
         Late Fusion & \textbf{61.5} & \textbf{48.2} & 63.6 & \textbf{71.4} & 55.8 & \textbf{55.6} & \textbf{40.5} & \textbf{49.0} & \textbf{16.9}\\ 
         \bottomrule
    \end{tabular*}
    \label{tab:main}
\end{table*}

\subsection{Dataset Statistics}
The dataset statistics are shown in Figure~\ref{fig:statistics}. The quantification of semantic voxels within each category over all scenes in the V2VSSC dataset is conducted, excluding voxels that are classified as empty. The graphic illustrates that the number of poles is the lowest, while the car count is slightly greater than that of poles. On the other hand, road, terrain, building, and vegetation exhibit the highest frequencies. This observation aligns with our intuitive perception of three-dimensional situations.

\section{Experiments}
\subsection{Benchmark models:}
\noindent\textbf{Metadata Sharing:} The first step involves transmitting the extrinsics and relative position of each CAV in order to construct a spatial graph. In this graph, each node represents an agent within the communication range, while each edge symbolizes a communication channel connecting a pair of nodes. The spatial map undergoes real-time updates while the vehicle is in operation. Nodes are removed from the spatial map when they exceed the communication range, and are added to the map when they enter the communication range. Subsequently, the nearby CAV communicate data with the ego vehicle based on the designated message sharing technique. Ultimately, the ego vehicle generates a semantic occupancy map by using feature extractors and detecting heads. 

\noindent\textbf{Feature Extraction:} We have used the most advanced LiDAR-based 3D object detectors, namely VoxelNet~\cite{zhou2018voxelnet}, in our dataset. Additionally we integrate it with three distinct fusion algorithms, namely early fusion, intermediate fusion and late fusion. we also examine the performance of the model in a scenario where only one vehicle is considered, referred to no fusion.

\noindent\textbf{Message Fusion Baseline:} 1) Early fusion baseline. The LiDAR point clouds will be transformed to the coordinate frame of the ego vehicle based on the posture information exchanged by CAVs in the spatial graph. Subsequently, the ego vehicle will aggregate all received point clounds and feed them to the detector. 2) Intermediate fusion baseline. Attentive modules are used as our intermediate pipeline. The LiDAR detector is used to extract features from each vehicle, after which the intermediate features are broadcasted. When ego vehicle receive the broadcast information, it will integrate its intermediate features using an attention mechanism. Ultimately, the detection head produces the ultimate semantic occupancy map. The structure of the Intermediate fusion baseline is shown in Figure.~\ref{fig:system}. This particular fusion baseline is chosen for illustration because of its higher complexity compared to the other two fusion methods. 3) Each CAV will provide a prediction of the 3D semantic occupancy map, including confidence scores for each occupancy grid, and transmit these outputs to the ego vehicle. Non-maximum suppression (NMS) is used to update the semantic occupancy map of the ego vehicle, guaranteeing that each occupancy grid retains the greatest confidence score.

\begin{figure*}[!t]
\centering
    \begin{subfigure}[c]{0.31\linewidth}
        \centering{\includegraphics[width=1\linewidth]{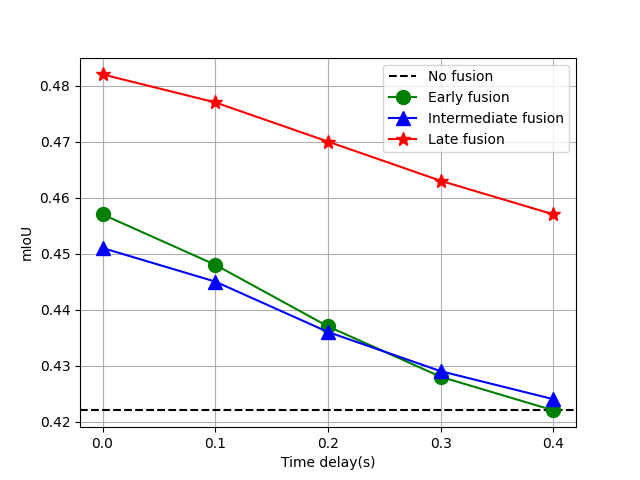}}

        \caption{Time delay(s)}
        \label{fig:output-a}
    \end{subfigure}
    \begin{subfigure}[c]{0.31\linewidth}
        \centering{\includegraphics[width=1\linewidth]{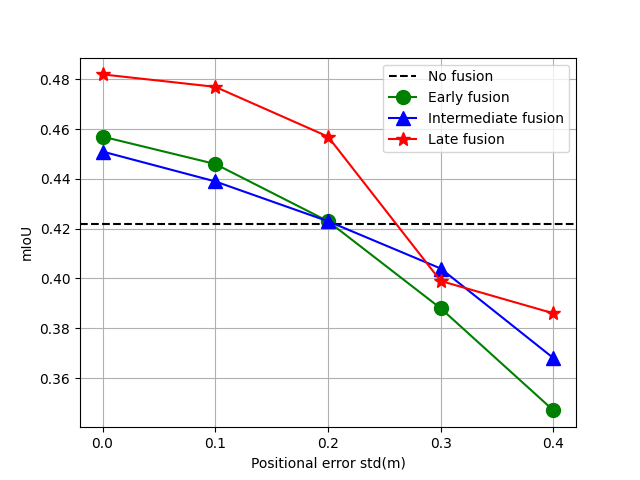}}
        \caption{Positional error std(m)}
        \label{fig:output-b}
    \end{subfigure}
    \vspace{1mm}
    \begin{subfigure}[c]{0.31\linewidth}
        \centering{\includegraphics[width=1\linewidth]{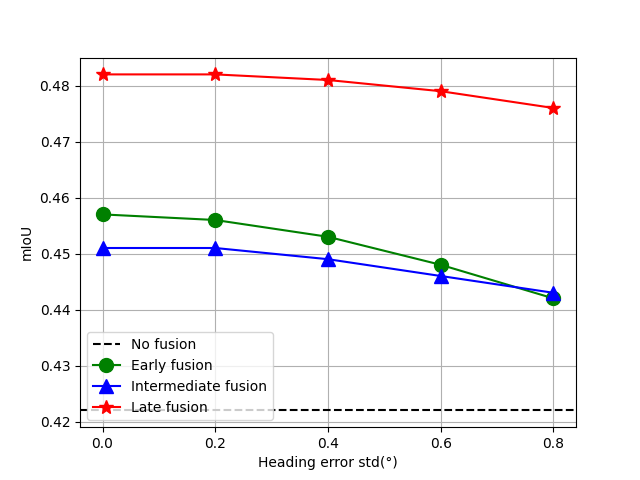}}
        \caption{Heading error std(°)}
        \label{fig:output-c}
    \end{subfigure}

    \caption{\textbf{Time delay and location error}}
    \label{fig:visda}
    \vspace{-2mm}
\end{figure*}

\subsection{Metrics:}
The evaluation of detection performance is conducted in close proximity to the ego vehicle within a specified range of distances. Specifically, the evaluation is performed within the range of $x \in [50, 50]m$, $y \in [50, 50]m$, $z \in [-3, 5]m$. The computational load of high-resolution 3D prediction is a significant factor contributing to its complexity~\cite{wang2023openoccupancy}. Efficiency and communication bandwidth are taken into consideration while determining the voxel resolution. Specifically, the voxel resolution is set as $0.78m$ for X,Y axis, and $0.4m$ for Z axis, resulting in a volume of $20 \times 128 \times 128$ voxels for occupancy prediction. For evalution metrics, we utilize Intersection of Union(IoU) as the geometric metric and the mean Iou(mIoU) for each class as the semantic metric following the SSCNet~\cite{song2017semantic}. In additional, we propose a critical objects Iou(cIoU) which was set as the most concerned drivable areas and vehicles for autonomous driving. IoU identifies a voxel as being occupied or empty, mIoU identifies a voxel as being classified correctly or not and cIoU directly reflects the most concerned performance for autonomous.

\subsection{Implementation details:}
The train/validation/test splits are 6764/1980/2170. We select a fixed vehicle as the ego vehicle among all spawned CAVs for each scenario in the validaton and test set. According to communication theory analysis, we set the spatial graph range to be 70 meters. Outside of this communication range will be ignored by the ego vehicle. We first train the model under the Perfect Setting which no noise and time delay. Then fine-tune the model for Noisy Setting. We adopt Adam optimizer~\cite{kingma2014adam} with an initial learning rate of $2 * 10^{-3}$, and employ cosine anneal warm restart schedule with initial learning rates of $2 * 10^{-4}$, $\eta_{min} = 2 * 10^{-5}$, and $T_0 = 10$. All models are trained on Tesla V100.

\subsection{Benchmark Analysis:}
Table~\ref{tab:main} displays the performance of VoxelNet paired with several fusion algorithms on both Perfect and Noisy Setting. All the fusion algorithms achieve IoU, mIoU and cIoU surpass No Fusion baseline. It can be observed from the table that the three fusion approaches are basically higher than no-fusion in all semantic categories under ideal circumstances, which indicates the superiority of cooperative perception in SSC tasks. Among them, the early fusion method produces the highest critical objects metric, which is 6.8\% higher than No-fusion, completely showing the benefits of raw data fusion. The intermediate fusion also displays superior critical objects metrics. Compared with the early fusion, it performs better for terrain, vegetation, and poles. The late fusion exhibits excellent performance indicators. Although it is inferior to the early fusion in vehicle prediction, it is significantly ahead of other methods in geometric metrics and semantic metrics. For no-fusion, it is raised by 8.3\% and 6.0\% correspondingly.

\subsection{Time delay analysis}
In this study, we want to examine the effects of a time delay range $[0, 400]ms$. As depicted in Figure.~\ref{fig:output-a}, the performance of the three methods exhibits a decline as the time delay increases. When the delay time reaches $400ms$, significantly surpassing the actual lag time, both the early fusion method and the middle fusion method experience a substantial decline in performance, approaching a state similar to no-fusion. Conversely, the late fusion method maintains a performance level approximately 3.5\% higher than that of no-fusion when the delay time reaches $400ms$, indicating the robustness of the late fusion method to the time delay.

\subsection{Sensitivity to localization error}
Positioning error is a prevalent issue that arises in the context of autonomous driving. We established varying levels of Positioning noise and heading noise for the three fusion methods in order to replicate the error in Positioning. The Gaussian distribution is employed for sampling noise, with the standard deviation of position noise set to a range of $[0, 0.4]m$ and heading noise set to a range of $[0, 0.8]$°. As seen in Figure.~\ref{fig:output-b}, it can be observed that the performance of the fusion methods mIoU surpasses that of the no-fusion method when the position error is maintained within the usual range~\cite{li2020toward}. However, when the level of positional noise is massive(e.g. $0.4m$ twice higher than the maximum normal error value), the performance of fusion methods experiences a quick reduction and falls below that of the no-fusion method. One potential explanation is that a significant positioning error can introduce semantic ambiguity throughout the three-dimensional space of the ego vehicle, leading to a rapid deterioration in performance. In contrast, as depicted in Figure.~\ref{fig:output-c}, the fusion methods exhibit strong robustness in terms of the inaccuracy caused by heading noise. Even when confronted with an error value that is twice the highest normal threshold~\cite{xia2021advancing}, the method exhibiting the greatest reduction among the three ways is merely 1.5\%, which indicates that the fusion algorithm is robust to heading noise.

\begin{figure}[!t]
    \centering
    \includegraphics[width=0.7\linewidth]{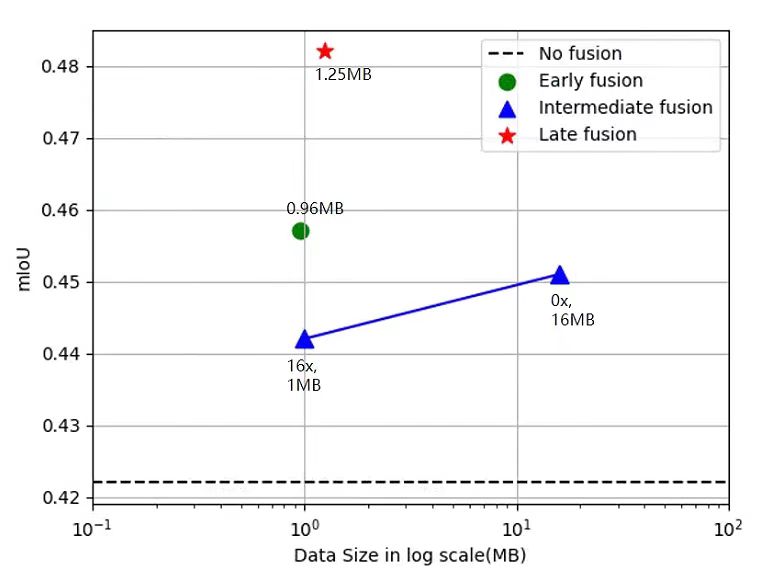}
    \caption{mIoU with respect to data size in log scale based on VoxelNet detector. The number $\times$ refers to the compression rate.}
    \label{fig:compress}
\end{figure}

\subsection{Effect of Compression Rates}
Figure.~\ref{fig:compress} exhibits the size of vehicle cooperative data and relative mIoU for all fusion methods on the testing set in V2VSSC dataset. Considering the bandwidth requirements, we investigate the effect of compressed intermediate features on performance and simulate the compression of intermediate features by modifying the number of layers by Encoder-Decoder. The performance of the mid-stage fusion method with 16$\times$ compression rate is 2\% better than no-fuison method which slightly drop (arround 1\%), but it can achieve millisecond latency with the early fusion and late fusion.

\section{CONCLUSIONS}
In this paper, we propose V2VSSC which is the first benchmark for collaborative semantic scene perception. The benchmark encompasses three distinct ways for vehicle-to-vehicle fusion and provides a V2VSSC dataset which serves as a valuable resource for facilitating multi-vehicle collaborative SSC tasks. The experimental results demonstrate that our system has the capability to partially address the challenges associated with single vehicle-induced difficulties in SSC tasks. In the future, it is our aspiration to validate practical concerns inside the realm of reality.

\addtolength{\textheight}{-3cm}   

\bibliographystyle{IEEEtran}
\bibliography{IEEEfull}
\end{document}